\documentclass[runningheads]{llncs}
\usepackage[T1]{fontenc}
\usepackage{graphicx}

\usepackage{cite}
\usepackage[pagebackref=true,breaklinks=true,letterpaper=true,colorlinks,bookmarks=false]{hyperref}
\usepackage{amsmath}
\usepackage{booktabs} 
\usepackage{multirow}
\usepackage{multicol}
\usepackage{tabularx,colortbl}
\usepackage{pifont}
\usepackage[dvipsnames]{xcolor}

\usepackage[ruled,vlined]{algorithm2e}

\usepackage{xspace}
\makeatletter
\DeclareRobustCommand\onedot{\futurelet\@let@token\@onedot}
\def\@onedot{\ifx\@let@token.\else.\null\fi\xspace}
\def\eg{\emph{e.g}\onedot} 
\def\ie{\emph{i.e}\onedot}

\usepackage{color}

\begin{document}

\definecolor{commentcolor}{RGB}{110,154,155}   
\newcommand{\PyComment}[1]{\ttfamily\textcolor{commentcolor}{\# #1}}  
\newcommand{\PyCode}[1]{\ttfamily\textcolor{black}{#1}} 

\title{Exploring Cycle Consistency Learning in Interactive Volume Segmentation}

\author{
Qin Liu\inst{1} \and
Meng Zheng\inst{2} \and
Benjamin Planche\inst{2} \and
Zhongpai Gao\inst{2} \and \\
Terrence Chen\inst{2} \and 
Marc Niethammer\inst{1} \and 
Ziyan Wu\inst{2}}

\institute{
University of North Carolina at Chapel Hill \and
United Imaging Intelligence, Cambridge MA, USA \\
\href{https://github.com/uncbiag/iSegFormer/tree/v2.0}{https://github.com/uncbiag/iSegFormer/tree/v2.0}
}

\maketitle

\begin{abstract}
Automatic medical volume segmentation often lacks clinical accuracy, necessitating further refinement. In this work, we interactively approach medical volume segmentation via two decoupled modules: \emph{interaction-to-segmentation} and \emph{segmentation propagation}. Given a medical volume, a user first segments a slice (or several slices) via the interaction module and then propagates the segmentation(s) to the remaining slices. The user may repeat this process multiple times until a sufficiently high volume segmentation quality is achieved. However, due to the lack of human correction during propagation, segmentation errors are prone to accumulate in the intermediate slices and may lead to sub-optimal performance. To alleviate this issue, we propose a simple yet effective cycle consistency loss that regularizes an intermediate segmentation by referencing the accurate segmentation in the starting slice. To this end, we introduce a backward segmentation path that propagates the intermediate segmentation back to the starting slice using the same propagation network. With cycle consistency training, the propagation network is better regularized than in standard forward-only training approaches. Evaluation results on challenging AbdomenCT-1K and OAI-ZIB datasets demonstrate the effectiveness of our method.

\end{abstract}

\section{Introduction}
Volumetric medical image segmentation is critical in various applications, ranging from disease diagnosis~\cite{shen2017deep,litjens2017survey,devunooru2021deep,heimann2010segmentation} to surgical planning and treatment~\cite{li2021medical,soler2001fully}. Automatic approaches often lack clinical accuracy, necessitating further corrections~\cite{wang2018interactive}. Interactive segmentation, which allows users to refine automated segmentations with additional hints (\eg, clicks~\cite{liu2022simpleclick}, scribbles~\cite{cheng2021modular}, and bounding boxes~\cite{zhang2020interactive}), appears to be an effective way for segmentation refinement. Therefore, we focus on interactive volume segmentation in this work.

Highly successful temporal propagation modules have been proposed for semi-supervised video object segmentation (VOS)~\cite{oh2019video,cheng2021rethinking,cheng2022xmem}. Inspired by these approaches, recent interactive volume segmentation methods~\cite{liu2022isegformer,zhou2023volumetric,shi2022hybrid} have used a modular framework that decouples human interaction from segmentation propagation. In this way, a video temporal propagation module can be directly applied to medical volumes. Conceptually, these modular methods combine two tasks: interaction-to-segmentation (\ie, interactive segmentation on 2D slices~\cite{liu2022simpleclick}) and segmentation propagation (\ie, as in temporal propagation in semi-supervised VOS~\cite{cheng2021rethinking}). These modular methods differ significantly from unified 3D approaches~\cite{diaz2022deepedit,liao2020iteratively}, which directly adopt 3D encoders and decoders for interactive volume segmentation. Modular approaches, as opposed to unified ones, allow users to focus on segmenting a single slice with high quality without having to check the interaction effects on other slices, as checking itself takes time and effort. Besides, modular methods offer greater flexibility in supporting various user interactions since interaction and propagation are decoupled. In the era of large foundation models, modular methods benefit from the advances of both modules~\cite{kirillov2023segment,cheng2023putting}. Thus, we focus on modular interactive volume segmentation.

The propagation module in existing modular approaches~\cite{zhou2023volumetric,shi2022hybrid,liu2022isegformer} rely on the state-of-the-art Space Time Memory network (STM)~\cite{oh2019video} and Space Time Correspondence Network (STCN)~\cite{cheng2021rethinking}). STM builds a memory bank that stores representations for intermediate images and their segmentations. A query image retrieves a segmentation from this memory bank by the learned correspondence between query and memory representations. STCN improves STM by introducing a much more memory-efficient correspondence learning approach. 
STM and STCN use a sequential propagation order to support online video processing. This online restriction is unnecessary in medical volumes, typically acquired offline before segmentation. Moreover, segmentation errors can accumulate in intermediate slices without human correction, potentially degrading the memory bank and leading to suboptimal performance.

To alleviate this issue, we propose a simple yet effective cycle consistency loss that regularizes an intermediate segmentation by referencing an accurate segmentation in the starting slice. To this end, we introduce a backward segmentation path that propagates the intermediate segmentation back to the starting slice using the same propagation network. Compared with the propagated segmentation of intermediate slices, the segmentation of the starting slice is always accurate and reliable. Therefore, the accurate segmentation in the starting slice will alleviate the segmentation errors of an intermediate slice that is flowing in a forward-backward loop. We evaluated cycle consistency training on several public benchmarks, including AbdomenCT-1K~\cite{Ma-2021-AbdomenCT-1K} and OAI-ZIB~\cite{ambellan2019automated}. 
Evaluation results on the AbdomenCT-1K dataset show that cycle consistency training improved segmentation on 8 out of 12 organs, including \textbf{24.9\%} and \textbf{8.3\%} improvements for the challenging esophagus and inferior vena cava, respectively. 

Our contributions are as follows: 1) we investigate the error accumulation problem of existing propagation modules for interactive volume segmentation and mitigate it by better training regularization; 2) we explore a simple yet effective cycle consistency training strategy that encourages self-correction for the intermediate propagated segmentations; and 3) we demonstrate the effectiveness of our proposed method on challenging and diverse benchmarks. To the best of our knowledge, we are the first to explore cycle consistency learning in interactive volume segmentation.

\section{Related Work}

\noindent\textbf{Video object segmentation (VOS).} We focus on semi-supervised VOS that aims to segment a video object given the segmentation of the first frame~\cite{gao2023deep}. We are particularly interested in approaches such as STM~\cite{oh2019video} and STCN~\cite{cheng2021rethinking} that build a memory bank for explicit pixel-level matching. STM was a breakthrough memory network that inspired many follow-up works such as SwiftNet~\cite{wang2021swiftnet} and STCN. Even though these memory networks were developed for videos, they have recently been applied for interactive segmentation on medical volumes~\cite{zhou2023volumetric,liu2022isegformer,shi2022hybrid}. However, all these methods have the inherent memory network limitation of accumulating propagation errors in an expanding memory bank. \emph{In this work, we explore using cycle consistency learning to mitigate this problem for interactive volume segmentation.} 

\noindent\textbf{Interactive volume segmentation.} Modular interactive volume segmentation approaches have been proposed in~\cite{zhou2023volumetric,liu2022isegformer,shi2022hybrid}. Mem3D~\cite{zhou2023volumetric} first applies STM~\cite{oh2019video} as the propagation network for interactive volume segmentation. It further proposes a quality assessment module to recommend the next slice for refinement. iSegFormer~\cite{liu2022isegformer} uses STCN~\cite{cheng2021rethinking}, an upgraded STM, as the propagation module and shows that the STCN model trained on videos can perform well on medical images even without finetuning. HybridNet~\cite{shi2022hybrid} proposes a hybrid propagation network that combines STCN with a 3D encoder-decoder network to provide spatially consistent features, demonstrating these features' effectiveness in improving interactive segmentation performance. \emph{In this work, we use STCN as the propagation network due to its strong performance. Unlike previous methods, we focus on exploring cycle-consistent learning in interactive volume segmentation.} 

\noindent\textbf{Cycle consistency learning.} Cycle consistency learning has been explored in various tasks such as video object segmentation~\cite{li2020delving}, image synthesis~\cite{zhu2017unpaired}, and unsupervised pretraining~\cite{dwibedi2019temporal,wang2019learning}. Besides, it has been applied to medical image segmentation~\cite{wang2022cycmis}, registration~\cite{kim2021cyclemorph,greer2021icon,tian2022gradicon}, and synthesis~\cite{wang2018unsupervised}. \emph{In this work, we explore cycle consistency learning in interactive volume segmentation. More specifically, we aim to address the error accumulation problem in the propagation module by regularizing intermediate segmentations via a cyclic mechanism.}  

\begin{figure}[h]
  \centering
  \includegraphics[width=12.0cm, height=6.0cm]{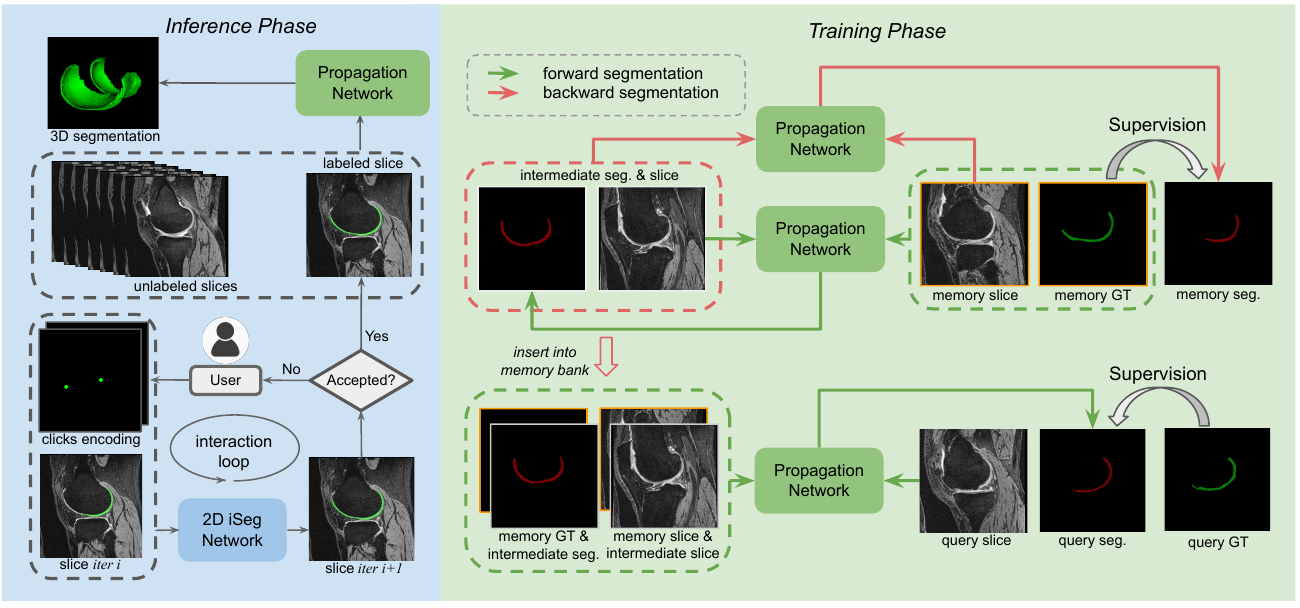}
  \caption{\emph{Method overview}. Left: our modular framework for interactive volume segmentation. Right: cycle-consistent training for the propagation module, in which all propagation networks share the weights. We introduce a backward segmentation path (shown in red arrows) into a standard training process that consists of only forward segmentation paths (shown in green arrows). ``GT'' denotes ground truth (shown in green); ``seg.'' denotes model segmentation (shown in red).}
  \label{fig:framework}
\end{figure}

\section{Cycle Consistency Learning}

\subsection{Problem Formulation}
\label{sec:problem_formulation}

Given a volume with $T$ slices, a memory slice $m$ is defined as the starting slice for propagation; a query slice $q$ is defined as the target slice to be propagated; an intermediate slice $p$ is defined as any slice that locates between $m$ and $q$. The ground truth segmentations of slices $m$, $p$, and $q$ are denoted as $m\_gt$, $p\_gt$, and $q\_gt$, respectively. A propagation network parameterized by $\theta$ is denoted as $net_\theta$, which segments the intermediate slice as below:
\begin{equation}\label{eq:eq1}
        p\_seg = net_\theta(p, [m, m\_gt]) 
\end{equation}
where $[\hspace{1pt}\cdot\hspace{1pt}]$ denotes a scalable memory bank that can be updated by appending more (slice, segmentation) pairs. For example, to segment the query slice, the memory bank may be updated as [$m$, $m\_gt$, $p$, $p\_seg$] by appending the intermediate slice $p$ and its segmentation $p\_seg$.

\begin{algorithm}[h]
\SetAlgoLined
    \PyComment{net: propagation network with learnable parameters} \\
    \PyComment{loss: segmentation loss function (\eg, cross-entropy loss)} \\
    \PyComment{lambda: loss weight} \\
    \PyComment{m, p, q: memory, intermediate, and query slices} \\
    \PyComment{m\_gt, q\_gt: ground truths for memory and query slices} \\
    \BlankLine
    \PyComment{forward segmentation} \\
    \PyCode{p\_seg, q\_seg = net(p, [m, m\_gt]), net(q, [m, m\_gt, p, p\_seg])} \\ 
    \BlankLine    
    \PyComment{backward segmentation} \\
    \PyCode{m\_seg = net(m, [p, p\_seg])} \\
    \BlankLine
    \PyComment{cycle loss} \\
    \PyCode{q\_loss, m\_loss = loss(q\_seg, q\_gt), loss(m\_seg, m\_gt)} \\
    \PyCode{cycle\_loss = q\_loss + lambda $*$ m\_loss}
\caption{Pseudocode for calculating cycle consistency loss}
\label{algo:your-algo}
\end{algorithm}

\subsection{Cycle Consistency Training} 
\label{sec:cycle_consistency_training}

For cycle consistency training, we randomly sample three slices (\ie a memory slice $m$, an intermediate slice $p$, and a query slice $q$ as defined in Sec.~\ref{sec:problem_formulation}). The ground truth segmentations of the three slices are also provided. Fig.~\ref{fig:framework} shows the process of training a propagation network $net_\theta$ (or $net$ in short) via the three slices. In a nutshell, cycle consistency training introduces a backward segmentation path (shown in red arrows) into standard training that only performs forward segmentation (shown in green arrows). Next, we will introduce the details of forward and backward segmentation.

\noindent\textbf{Forward segmentation.} There are two stages in forward segmentation: 1) the first stage propagates the memory ground truth $m\_gt$ to the intermediate slice $p$, resulting in an intermediate segmentation $p\_seg$. This stage is denoted as $p\_seg = net(p, [m, m\_gt])$, in which the memory bank $[m, m\_gt]$ only contains a memory slice and its ground truth. 2) the query slice then retrieves a segmentation via an updated memory bank $[m, m\_gt, p, p\_seg]$, which leverages the intermediate segmentation. This stage is denoted as $q\_seg = net(q, [m, m\_gt, p, p\_seg])$. 

\noindent\textbf{Backward segmentation.} To further regularize the intermediate segmentation $p\_seg$ obtained in forward segmentation, we propagate the intermediate segmentation $p\_seg$ back to the memory slice $m$ using the same propagation network $net$. Thus, we obtain $m\_seg$ and complete a propagation loop. The backward segmentation path can be denoted as $m\_seg = net(m, [p, p\_seg])$.

\subsection{Cycle Consistency Loss}
\label{lab:cycle_consistency_loss}

Alg.~\ref{algo:your-algo} shows the pseudo-code for obtaining the cycle consistency loss. Given the segmentations from the forward and backward paths (described in Sec.~\ref{sec:cycle_consistency_training}), we define cycle consistency loss $\mathcal{L}_{cycle}$ as 

\begin{equation}\label{eq:cycle_loss_definition}
    \begin{aligned}
        \mathcal{L}_{cycle} = \mathcal{L}(q\_seg, q\_gt) + \lambda \mathcal{L}(m\_seg, m\_gt)
    \end{aligned}
\end{equation}
where $\mathcal{L}$ denotes a conventional segmentation loss and $\lambda\geq 0$ is a weight for the memory segmentation loss. The query segmentation $q\_seg$ and the memory segmentation $m\_seg$ are supervised by their corresponding ground truths using the same loss function $\mathcal{L}$. The intermediate segmentation $p\_seg$ can also be supervised by its ground truth, but we ignore this loss in Eq.~\ref{eq:cycle_loss_definition} for brevity. When $\lambda = 0$, our cycle consistency loss degrades to the standard loss. 

\subsection{Implementation Details}
\label{lab:implementation_details}

\noindent\textbf{Networks.} For the propagation network, we use a video-pretrained STCN~\cite{cheng2021rethinking} as the baseline and fine-tune it on medical images with or without the proposed cycle consistency loss. STCN consists of a key encoder and a value encoder using ResNet50 and ResNet18, respectively. For the 2D interactive segmentation network, any existing open-source method~\cite{liu2022simpleclick,kirillov2023segment,zou2024segment} can be combined with STCN for interactive volume segmentation. This network is not the focus of our work and is only used in qualitative evaluation (see demos in the appendix).

\noindent\textbf{Training.}
We set $\lambda=0.1$ in the cycle consistency loss based on parameter tuning on video datasets (Sec.~\ref{sec:ablations}). We adopt most of the hyperparameters introduced in STCN~\cite{cheng2021rethinking} for a fair comparison with the baseline. For example, we follow the same data augmentation strategy that randomly reverses the order of the slices in a training batch; we also use bootstrapped cross entropy as the segmentation loss in Alg.~\ref{algo:your-algo} and Eq.~\ref{eq:cycle_loss_definition}. Our implementation is based on Python and PyTorch. Our models are trained and evaluated on an NVIDIA RTX A6000 GPU. We fix the number of finetuning iterations to 10k without model selection.

\section{Experiments}

\subsection{Benchmarks}
\label{lab:experiments}

\noindent\textbf{Datasets.} We evaluate our method on two datasets: AbdomenCT-1K~\cite{Ma-2021-AbdomenCT-1K} and  OAI-ZIB~\cite{ambellan2019automated}. \textbf{AbdomenCT-1K} contains 50 CT abdomen volumes (4794 slices) with manual segmentations for 12 diversified organs (Fig.~\ref{fig:abd1k-cycle-relative-improvement}), including tiny and challenging organs such as the pancreas and esophagus, as well as large and regular organs such as the liver and kidney. Our experiments leverage large-scale, video-pretrained weights for the propagation model, requiring minimal data for fine-tuning and allocating the bulk of the data for evaluation. This experiment setting mirrors the scarcity of annotated medical volumes. We randomly partitioned the AbdomenCT-1K dataset into 10 volumes (955 slices) for fine-tuning and 40 volumes (3839 slices) for testing, utilizing all 12 organs in both phases. Consequently, these organs are classified as ``seen'' organs. We also generate rib segmentations for the testing volumes to evaluate the model's generalizability and use them as ``unseen'' organs. We generate rib segmentations using our internally developed rib segmentation model with minor manual refinement (an example is shown in the appendix). \textbf{OAI-ZIB} consists of 507 3D MR images with segmentations for the femur, tibia, tibial cartilage, and femoral cartilage. In this work, we only conduct a qualitative evaluation using this dataset. 

\noindent\textbf{Evaluation protocol.} For \emph{quantitative} evaluation, we only conduct one round of propagation for simplicity and efficiency; we evaluate each organ separately and use the ground truth for propagation to decouple the interaction module. We follow the same propagation strategy introduced in STCN~\cite{cheng2021rethinking}. For \emph{qualitative} evaluation, we conduct multiple rounds of propagation and rely on the interaction module to provide the first segmentation and refine propagation results.

\noindent\textbf{Evaluation metrics.} We report Region Jaccard ($\mathcal{J}$), Boundary F measure ($\mathcal{F}$), and their average ($\mathcal{J\&F}$) to assess segmentation quality. We also report the Dice Similarity Coefficient (DSC) as an additional metric.

\begin{table}
\footnotesize
\centering
\begin{tabular}{l c c c c c c c c c c c c}
    \toprule
    & \multicolumn{4}{c}{AbdomenCT-1K} & \multicolumn{4}{c}{Left Rib (unseen)} & \multicolumn{4}{c}{Right Rib (unseen)}\\
    \cmidrule(lr){2-5} \cmidrule(lr){6-9} \cmidrule(lr){10-13}
    & $\mathcal{J}$ & $\mathcal{F}$ &  $\mathcal{J\&F}$ & $DSC$ & $\mathcal{J}$ & $\mathcal{F}$ &  $\mathcal{J\&F}$ & $DSC$ & $\mathcal{J}$ & $\mathcal{F}$ &  $\mathcal{J\&F}$ & $DSC$ \\    
    \midrule
    STM~\cite{oh2019video} & 31.4 & 39.8 & 35.6 & 47.8 \\
    STCN~\cite{cheng2021rethinking} & 56.0 & 74.3 & 66.1 & 71.8 & 27.4 & 45.2 & 36.3 & 43.0 & 23.3 & 36.9 & 30.1 & 37.8 \\
    \rowcolor[gray]{0.95}    
    +FT w/o cycle                   & 64.9 & 82.4 & 74.7 & 78.7 & 38.3 & 59.0 & 48.7 & 55.4 & 33.9 & 57.0 & 45.4 & 50.6 \\
    \rowcolor[gray]{0.95}    
    +FT w/ cycle                    & \textbf{66.8} & \textbf{85.1} & \textbf{76.9} & \textbf{80.1} 
                                    & \textbf{41.0} & \textbf{64.6} & \textbf{52.8} & \textbf{58.2} 
                                    & \textbf{35.6} & \textbf{60.7} & \textbf{48.2} & \textbf{52.5} \\
    \bottomrule
\end{tabular}
\caption{\emph{Quantitative comparisons of our method and baselines on AbdomenCT-1K~\cite{Ma-2021-AbdomenCT-1K}}. We observe significant improvement when finetuning the SOTA video-pretrained STCN model on medical images with our proposed cycle consistency loss.}
\label{tab:main_results_abdomen1k}
\end{table}

\subsection{Comparisons}
\label{sec:main-results}

\noindent\textbf{Baselines.} We use two popular video object segmentation models, STM~\cite{oh2019video} and STCN~\cite{cheng2021rethinking}, as the baselines. The baselines are pretrained on videos. We finetune STCN on medical datasets w/ and w/o cycle consistency loss. See Tab.~\ref{tab:main_results_abdomen1k} for comparisons.

\begin{figure*}
  \centering
  \includegraphics[width=11.5cm, height=3.0cm]{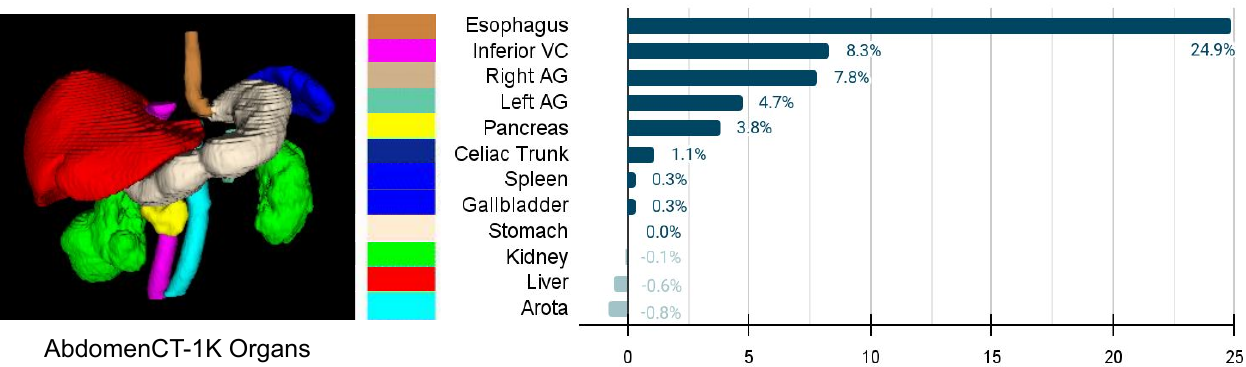}
  \caption{Relative $\mathcal{J\&F}$ improvement on the AbdomenCT-1K dataset. Although the performance drops slightly for 3 organs, we observe improvements on 8 out of the full set of 12 organs, resulting in average improvements across all metrics (Tab.~\ref{tab:main_results_abdomen1k}).}
  \label{fig:abd1k-cycle-relative-improvement}
\end{figure*}

\noindent\textbf{Quantitative comparisons.} Tab.~\ref{tab:main_results_abdomen1k} shows quantitative comparison results on the AbdomenCT-1K dataset. We compare three models: 1) a baseline STCN model that was pretrained on videos without being finetuned on AbdomenCT-1K; 2) a finetuned baseline model without the cycle consistency learning; 3) a finetuned baseline model with cycle consistency learning. In addition to the 12 labeled organs (\ie, ``seen'' organs), we generate the segmentations of ribs (\ie, ``unseen'' organs) for the testing volumes to evaluate a model's generalizability. All results in Tab.~\ref{tab:main_results_abdomen1k} are averaged over all volumes and all organs. First, we observe that finetuning an STCN model on medical images can significantly improve its performance. Second, our results show that cycle consistency learning can further boost the model's performance, for both seen and unseen organs, demonstrating the benefits of our method. The relative improvement for each organ is shown in Fig.~\ref{fig:abd1k-cycle-relative-improvement}.

\begin{figure}
  \centering
  \includegraphics[width=12.0cm, height=5.0cm]{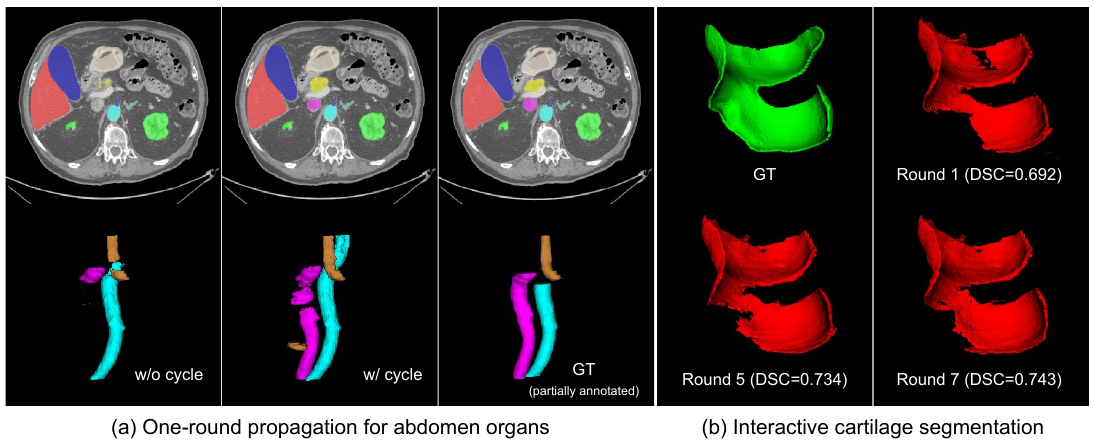}
  \caption{\emph{Qualitative comparison of our method and baseline on AbdomenCT-1K~\cite{Ma-2021-AbdomenCT-1K} and OAI-ZIB~\cite{ambellan2019automated}}. With cycle consistency learning, the STCN model achieves better performance for abdomen vessel segmentation.}
  \label{fig:qualitative-results}
\end{figure}

\noindent\textbf{Qualitative comparisons.} We compare our method (FT w/ cycle) with our baseline (FT w/o cycle) in Fig.~\ref{fig:qualitative-results}. As the ground truth for the Aorta (light blue) is not fully annotated at the top, our cycle consistency loss shows lower performance on the Aorta (Fig.~\ref{fig:abd1k-cycle-relative-improvement}) because it extends it beyond the annotation.

\subsection{Ablations}
\label{sec:ablations}

We performed ablations on video datasets to obtain the best $\lambda$. We finetuned an STCN baseline on the training set of YouTubeVOS 2018~\cite{xu2018youtube} with different values for $\lambda$. Evaluation results of the validation set show that $\lambda=0.1$ achieves the best performance; this is the value we used for our experiments in Tab.~\ref{tab:main_results_abdomen1k}.

\begin{figure}
  \centering
  \includegraphics[width=11.0cm, height=3.6cm]{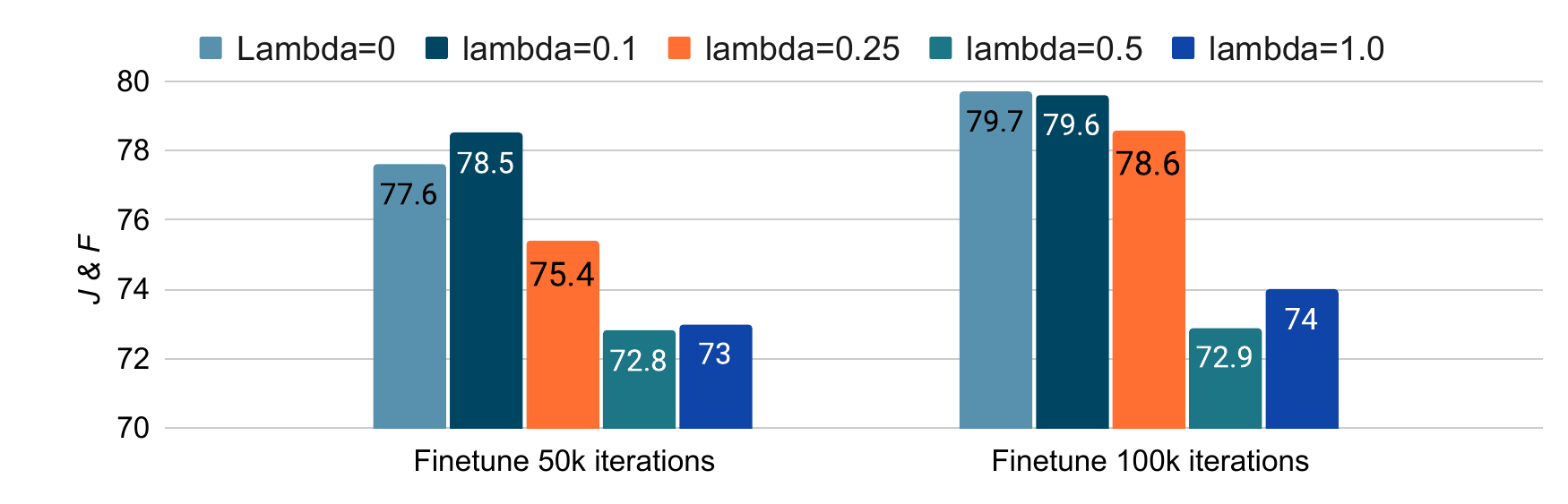}
  \caption{Ablation study on YouTubeVOS 2018~\cite{xu2018youtube}. $\lambda=0.1$ achieves the best result.}
  \label{fig:ablations}
\end{figure}

\vspace{-10pt}
\section{Conclusion}

We explored cycle consistency learning for interactive volume segmentation, aiming to address the pervasive error accumulation issue plaguing propagation modules. By introducing a segmentation backward path and integrating a cycle consistency loss, we seamlessly wove these advancements into current methodologies for medical volume segmentation. The efficacy of our strategy was rigorously affirmed through evaluations on challenging benchmarks, paving the road for developing more capable and advanced methods in the field.

{\small
\bibliographystyle{ieeetr}
\bibliography{main}
}

\end{document}